\documentclass[lettersize,journal, twoside]{IEEEtran}  

\usepackage{amsmath,amsfonts}
\usepackage{array}
\usepackage[caption=false,font=normalsize,labelfont=sf,textfont=sf]{subfig}
\usepackage{textcomp}
\usepackage{stfloats}
\usepackage{url}
\usepackage{verbatim}
\usepackage{graphicx}
\usepackage{balance}
\usepackage{algorithm}
\usepackage{algpseudocode}
\usepackage{booktabs}
\usepackage{multirow}
\usepackage{tabularx}
\usepackage{xcolor}  

\usepackage[colorlinks=true, linkcolor={rgb:red,2;green,1;blue,0.5}, citecolor=blue, urlcolor=red]{hyperref}

\title{\LARGE \bf
DynaWM: Dynamics-Aware Distillation with World Model and Momentum Targets for Smooth Locomotion over Continuous Stairs
}

\author{Haidong Hou, Zhangguo Yu, Hengbo Qi, and Jianlin Zhang 
\thanks{This work was supported in part by the National Natural Science Foundation of China under Grant 52575004, and in part by the Beijing Natural Science Foundation under Grants L252015 and L243004.}
\thanks{The authors are with the School of Mechatronical Engineering, Beijing Institute of Technology, Beijing 100081, China (e-mail: houhaidong@bit.edu.cn; corresponding author: yuzg@bit.edu.cn).}%
}

\begin{document}

\maketitle
\markboth{IEEE/RSJ International Conference on Intelligent Robots and Systems (IROS)}%
{Hou \MakeLowercase{\textit{et al.}}: DynaWM: Dynamics-Aware Distillation with World Model and Momentum Targets}

\begin{abstract}

Recent advances in control have enabled bipedal-wheeled robots to traverse slopes and single-step obstacles, yet long staircase traversal remains challenging as current teacher-student frameworks suffer from weakened dynamics-aware representations and incomplete terrain geometry encoding. To bridge this gap, we propose DynaWM, a dynamics-aware representation learning framework. To enhance terrain encoding capability and enable transparent assessment, we introduce a world model as a regularizer to enforce forward-dynamics awareness, preserving comprehensive terrain geometry while facilitating hierarchical encoding visualization. To stabilize knowledge transfer, we employ a momentum target encoder to provide consistent distillation targets, preventing dimensional collapse from non-stationary teacher updates. Evaluation of the learned representations through Principal Component Analysis (PCA) visualization and quantitative metrics reveals that our encoder hierarchically captures terrain geometry with higher terrain encoding capability, leading to enhanced terrain adaptability and motion smoothness. Experimental results in simulation and real hardware demonstrate that our method achieves superior terrain adaptability and motion smoothness, enabling bipedal-wheeled robots to overcome diverse continuous stairs, as shown in Fig. \ref{first_picture}.

\end{abstract}

\section{INTRODUCTION}

\par By combining the energy efficiency of wheeled locomotion with the terrain adaptability of legged systems, bipedal-wheeled robots can cruise efficiently on flat surfaces while still being able to overcome obstacles by actively deploying their legs. They are now recognized as a promising platform for mobile robotics due to their hybrid design \cite{gao_multimode_2024}. Recent advances based on model predictive control \cite{klemm_ascento_2019,zhao_compliant_2024} and deep reinforcement learning \cite{chamorro_reinforcement_2024,yang_multi-loco_2025} have enabled these platforms to successfully traverse slopes, grasslands, and single-step obstacles, demonstrating substantial potential for locomotion in complex terrain and gradual deployment in practical applications such as logistics delivery and industrial inspection.
\par To further extend robot capabilities to complex terrains such as continuous stairs, accurate terrain perception and adaptation mechanisms are essential. However, the sim-to-real gap \cite{TWIST} and hardware unreliability \cite{sun2025_learning_perceptive} make direct integration of exteroceptive perception modules extremely difficult. To address these challenges, privileged learning paradigms, particularly teacher-student distillation frameworks, have emerged as the dominant solution for incorporating terrain awareness into control policies \cite{kim2024NotOnlyRewards}. These approaches first train a privileged teacher policy with access to exteroceptive information, such as terrain elevation maps and contact forces, and subsequently adopt DAgger \cite{2011_DAgger} distillation to transfer knowledge to a student policy relying solely on proprioceptive observations. By circumventing the challenges of direct perception integration, such as sensor unreliability and sim-to-real gaps, these methods enable the student to retain terrain awareness without requiring explicit exteroceptive inputs during deployment \cite{2025_Unitracker}.

\begin{figure}[t]
   \centering
   \includegraphics[scale=0.29]{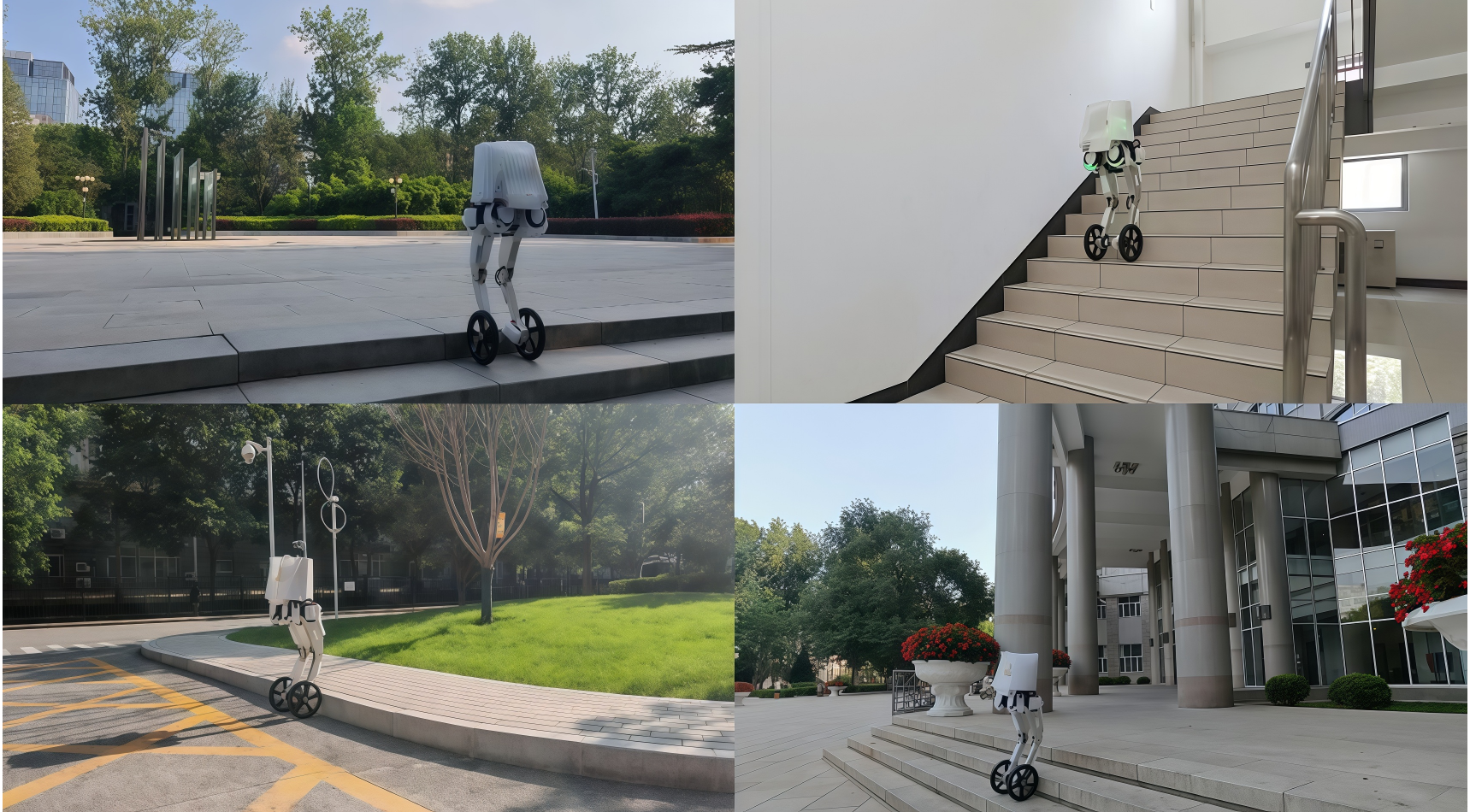}
   \caption{Models trained using our method were evaluated across diverse 
   outdoor stairs of varying widths and heights.}
   \label{first_picture}
\end{figure}

\par Recent extensions have further enhanced this approach. \cite{kumar2021rma} introduced an environment encoder that enhances information retention and minimizes uncertainty by shifting the distillation target from raw observations to latent representations. Concurrent teacher-student training schemes, as proposed by \cite{Wang2024CTS}, enable the simultaneous training of both networks, significantly enhancing both training effectiveness and efficiency. Extensive validation on both quadrupedal and humanoid platforms has shown the effectiveness of these teacher-student frameworks in traversing challenging terrain, as they have adeptly traversed long staircases, slopes, and irregular surfaces.

\par However, there are three main issues with current approaches. First, teacher encoders optimized exclusively through policy gradients tend to weaken dynamics-aware representations \cite{whitney2020dynamics}. Specifically, they encode only terrain features directly relevant to immediate rewards while disregarding broader geometric information \cite{kumar2021rma}, leading to partial loss of terrain characteristics essential for ground dynamics modeling. Second, since the teacher encoder and student encoder are updated simultaneously, rapid changes in the teacher's dynamics-aware representations can cause dimensional collapse in the student encoder \cite{2025_represent_collapse,2025_Task-recency}. The student encoder tends to fit the teacher's mean predictions \cite{gao2025robust}, which can result in dimensional collapse and failure to capture essential terrain features. Third, the latent space is an opaque black box \cite{2021_visual_planning} where the true modeling capability for terrain height cannot be verified due to the interpretability issues in the training process. As a result, current approaches struggle to achieve dependable traversal over lengthy consecutive staircases due to their lack of motion smoothness and terrain adaptability when facing continuous stairs.

\par To address these limitations, we propose DynaWM (Dynamics-aware distillation with World model and Momentum targets), a dynamics-aware representation learning framework for bipedal-wheeled robot locomotion. By incorporating a world model \cite{2025_track_any_position} as a regularizer and a momentum target encoder for student distillation, our method significantly enhances the terrain encoding capability of encoders, leading to improved adaptability in continuous stairs traversal. Our contributions can be summarized as follows:
\begin{itemize}
\item We introduce a world model as a regularizer to enforce forward-dynamics awareness, which prevents the teacher encoder from disregarding geometric information unrelated to immediate rewards and ensures that the learned representations capture terrain dynamics rather than solely maximizing rewards.
\item We employ a momentum target encoder to provide stable distillation targets. By maintaining an exponential moving average of the student encoder, this approach prevents dimensional collapse caused by rapid changes in teacher representations during concurrent training.
\item We evaluate the learned terrain representations of encoders using PCA \cite{1901_0n_lines} visualization along with evaluation metrics, confirming that they encode terrain height in a hierarchical manner. As demonstrated in Fig.~\ref{first_picture}, these enhanced representations enable successful deployment in the real world, with robot overcoming continuous stairs through smoother motion and better terrain adaptation.
\end{itemize}


\section{Background}

\par Our framework builds upon reinforcement learning (RL) and Concurrent Teacher-Student (CTS) \cite{Wang2024CTS} distillation, enhanced by representation analysis techniques. We first formalize RL as a partially observable Markov decision process (POMDP), then review the CTS paradigm for simultaneous teacher-student training, and finally introduce PCA for
transparent assessment of learned representations.

\par RL is defined by the tuple $\langle \mathcal{S}, \mathcal{A}, \mathcal{P}, \mathcal{R}, \gamma \rangle$. Here, $\mathcal{S}$ denotes the state space, $\mathcal{A}$ the action space, $\mathcal{P}(s_{t+1} \mid s_t, a_t)$ the state transition probability, $\mathcal{R}(s_t, a_t)$ the immediate reward function, and $\gamma \in [0,1]$ the discount factor. The objective is to learn an optimal policy $\pi^*: \mathcal{S} \rightarrow \mathcal{A}$ that maximizes the expected cumulative discounted return:
\begin{equation}
J(\pi) = \mathbb{E}_{\pi} \left[ \sum_{t=0}^{\infty} \gamma^t R_t(s_t, a_t) \right].
\end{equation}

\subsection{Cocurrent Teacher-Student} Concurrent Teacher-Student (CTS) \cite{Wang2024CTS} is a reinforcement learning framework that trains teacher and student policies simultaneously within a single process. In CTS, all agents are split into teacher and student groups, and the collected data are correspondingly divided into two separate buffers, distinguished by superscripts ${i}\in\left\{t,s\right\}$. The teacher encoder and student encoder are trained on their respective datasets $\mathcal{D}^{t}$ and $\mathcal{D}^{s}$, producing latent representations $z^t_t$ and $z^s_t$. The teacher encoder and its associated actor network are optimized using Proximal Policy Optimization (PPO). The PPO loss for group $i$ is defined as:

\begin{equation}
\begin{aligned}
L^{{ppo},i}(\vartheta) =& \frac{1}{\mathcal{D}^i T} \sum_{\tau \in \mathcal{D}^i} \sum_{t=0}^{T}  \\
&\min \Big( r_t^i \hat{A}_t^i, \mathrm{clip}(r_t^i, 1+\epsilon, 1-\epsilon) \hat{A}_t^i \Big)
\end{aligned}
\end{equation}

\noindent where $r^i_t$ is the importance sampling ratio, $\hat{A}^i_t$ is the advantage estimated by Generalized Advantage Estimation (GAE), and 
$\vartheta$ denotes the combined parameters of the teacher encoder $\theta^t$ and the actor network $\theta$. This concurrent training paradigm enables both policies to be updated concurrently using reinforcement learning objectives, addressing the limitations of traditional two-stage teacher-student distillation.

The critic network is updated by minimizing the Mean Squared Error (MSE) between its predicted value $\widehat{V_t}$ and the target value $R_t$ estimated from the trajectory return.

\begin{equation}
    L^{value}\left(\phi\right)=\frac{1}{\mathcal{D}^tT}\sum_{\tau\in \mathcal{D}^t}\sum_{t=0}^{T}\left(\widehat{V}_t-R_t\right)^2
\end{equation}

\subsection{Principal Component Analysis} Principal Component Analysis (PCA) \cite{1901_0n_lines} is a classical linear dimensionality reduction technique that transforms high-dimensional data into a lower-dimensional space via orthogonal projection, where the resulting dimensions (principal components) capture the maximum variance from the original data. Given a centered data matrix $\mathbf{X}\in\mathbb{R}^{n\times d}$, PCA finds an orthogonal basis $\mathbf{W}\in\mathbb{R}^{d\times k}$ by solving $\max_{\mathbf{W}^{\top}\mathbf{W}=\mathbf{I}} \operatorname{tr}\left(\mathbf{W}^{\top}\mathbf{X}^{\top}\mathbf{X}\mathbf{W}\right)$, whose optimal solution is the eigenvectors corresponding to the largest eigenvalues of the covariance matrix $\mathbf{X}^{\top}\mathbf{X}$. 
\par In practice, a PCA plot refers to the scatter plot of data projected onto the first two or three principal components, enabling intuitive visualization of clustering structures, outliers, and latent patterns in high-dimensional data \cite{2025_field_performance}. In this work, we further compute the Pearson correlation coefficient between the first principal component scores and terrain height to quantify how effectively the encoder's primary variation direction captures height information \cite{2025_foods}.

\begin{figure*}[t] 
  \centering
  \includegraphics[width=\textwidth]{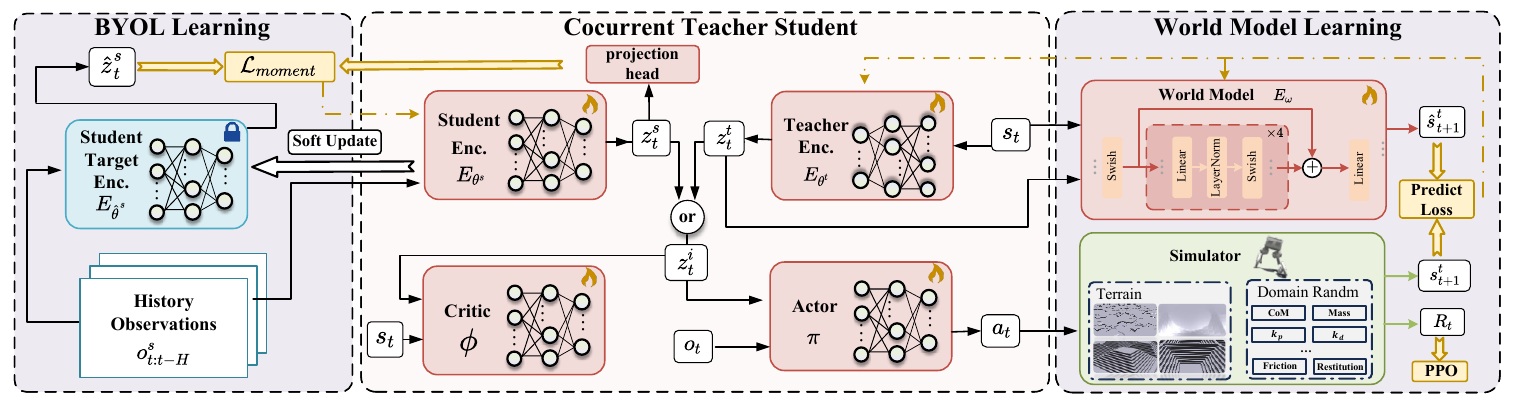} 
  \caption{\textbf{The DynaWM framework} comprises three interconnected modules. World Model Learning (right) regularizes the teacher encoder $E_{\theta^t}$ via state prediction loss $\mathcal{L}_{pred}$, enforcing forward-dynamics awareness. Concurrent Teacher-Student (center) processes privileged observations $s^t_t$ through the teacher encoder to latent $z^t_t$ for policy learning, while the student encoder $E_{\theta^s}$ learns from proprioceptive history $o^s_{t:t-H}$ to produce deployment-ready $z^s_t$ . BYOL Learning (left) employs a momentum target encoder $E_{\hat{\theta}^s}$ with soft update to provide stable distillation targets, where $\mathcal{L}_{moment}$ aligns the online student prediction to the target output, preventing dimensional collapse from teacher non-stationarity.}
  \label{framework}
\end{figure*}

\newcommand{\LongState}[1]{%
  \State \parbox[t]{\dimexpr\linewidth-\algorithmicindent}{#1\strut}%
}

\section{Method}

Our goal is to improve the terrain encoding capability of the Concurrent Teacher-Student (CTS) framework for bipedal-wheeled robots to enhance adaptability and smoothness in continuous stair traversal. To achieve this goal, we propose DynaWM, which incorporates a world model to enforce forward-dynamics awareness through state prediction and a momentum target encoder to provide stable distillation targets for student training, as shown in \ref{framework}. The remainder of this section details the world model architecture and the momentum-based distillation mechanism.

\subsection{World Model for Dynamics-Aware Representation}
To learn a dynamics-aware representation that encodes how terrain properties evolve and shape future robot states, the teacher encoder $E_{\theta^t}$ is designed to fulfill two crucial functions. The first function is to generate latent variables $z^t_t = E_{\theta^t}(s^t_t)$ for accurate forward state prediction. The second function is to supply terrain features to the policy network in order to facilitate optimal action selection. In order to achieve the primary goal, we present a world model that is capable of learning to forecast future terrain states based on the output from the encoder. The resulting prediction loss acts as a dynamics-aware regularizer, compelling the encoder to retain information essential for forward dynamics while discarding task-irrelevant noise. Below, we first detail the training mechanism that enables this dynamics-aware regularization, followed by the architectural design that ensures effective modeling of terrain dynamics.
\par \textbf{1) World Model Training}
\par The world model $W_\omega$ takes the teacher encoder's output $z^t_t$ together with the current privileged observation $s_t$ to predicts the next privileged state:

\begin{equation}
    \hat{s}^t_{t+1}=W_{\omega}\left(z_t^{t}, s^t_{t}\right)
\end{equation}

To train the world model and simultaneously propagate dynamics-aware gradients to the teacher encoder, we fix the actor–critic parameters and minimize the prediction residual:

\begin{equation}
    \min _{\theta^t, \omega} \mathcal{L}_{{pred }}(\theta^t, \omega)=\mathbb{E}_{\left(s_{t}, s_{t+1}\right) \sim \mathcal{B}}\left[\left\|W_{\omega}\left(z_t^{t}, s^t_{t}\right)-s_{t+1}\right\|^{2}\right]
\end{equation}

\noindent The gradient $\nabla_{\theta^t}\mathcal{L}_{{pred}}$ flows back to the teacher encoder, compelling it to extract essential dynamical variables while filtering out static or irrelevant features. This prediction loss serves as a dynamics-aware regularizer, ensuring the latent space remains grounded in actual terrain dynamics rather than merely maximizing immediate rewards.

\par \textbf{2) World Model Architecture}
\par We implement the world model as a deep residual network \cite{2017_resnet}. Each residual block follows:

\begin{equation}
\mathbf{h}_{i+1} = \mathbf{h}_i + \mathcal{F}_i(\mathbf{h}_i)
\end{equation}

\noindent where $\mathbf{h}_i$  is the input representation to the $i$-th block, $\mathbf{h}_{i+1}$ is the output representation,  $\mathcal{F}_i(\mathbf{h}_i)$ denotes the transformation learned through four successive Dense-LayerNorm-Swish units within the residual block. This residual design enables the network to model complex terrain dynamics through deep hierarchical feature extraction, rather than being limited to shallow input-output mappings as in typical MLPs. The skip connections facilitate stable gradient flow across many layers, allowing the world model to capture long-range temporal dependencies in terrain sequences and predict future states with higher accuracy. Consequently, the teacher encoder receives more informative forward-dynamics signals, guiding it toward representations that encode terrain geometry in a structured, interpretable manner.





\subsection{Momentum Target Encoder for Stable Distillation}
We introduce a momentum target encoder for student training, inspired by BYOL \cite{2021_BYOL}, to facilitate stable transfer of terrain-aware representations from the privileged teacher encoder to a student encoder, which relies solely on proprioceptive history $o^s_{t:t-H}$. The primary challenge lies in the non-stationary characteristics of the teacher encoder: its rapid updates during alternating optimization with the world model cause training instability and dimensional collapse \cite{2021_BYOL} when the student directly regresses to its outputs. To address this, we employ a momentum target encoder that maintains an exponential moving average of the student encoder, providing consistent distillation targets that evolve smoothly despite the teacher's fluctuations.

In particular, we maintain two distinct sets of student encoder parameters: the momentum target network $E_{\hat{\theta}^s}$, which employs an exponential moving average to gradually follow the online network, and the online network $E_{\theta^s}$, which is adjusted through gradient descent:

\begin{equation}
    \hat{\theta}_{t+1}^{s} = \tau \hat{\theta}_{t}^{s}+(1-\tau) \theta_{t}^{s}
\end{equation}

\noindent where $\tau \in [0.9,0.99]$ is a decay coefficient close to unity, and $(\cdot)_{t}$ denotes the value at time step $t$. The key insight is that the target encoder provides stable distillation targets that smooth out the teacher's rapid fluctuations, enabling robust student learning under non-stationary supervision.

Formally, let $\hat{z}^s_t=E_{\hat{\theta}^s}(o_{t:t-H}^s)$   denote the latent variable from the momentum target network, and $z^s_t=E_{\theta^s}(o_{t:t-H}^s)$ denote the online student latent variable. Following the BYOL \cite{2021_BYOL} paradigm, we apply a predictor head $q_{\xi}$ on top of the online network to introduce asymmetry between the online and target pathways, preventing the online encoder from collapsing to trivial representations identical to the target encoder \cite{2021_BYOL}. The distillation objective employs a normalized mean squared error loss:

\begin{equation}
\mathcal{L}_{{moment}} = \left\| \frac{q_{\xi}(z_t^s)}{\left\|q_{\xi}(z_t^s)\right\|_2} - \frac{\hat{z}_t^s}{\left\|\hat{z}_t^s\right\|_2} \right\|_2^2
\end{equation}

This loss pulls the online representation toward the normalized target representation. By aligning to the target encoder's output rather than directly to the teacher, the online student learns from stable targets that are insulated from the teacher's non-stationarity, effectively preventing dimensional collapse.

To ensure the target encoder accurately tracks the teacher's representation space, we optionally add a slow-updating alignment loss:
\begin{equation}
\mathcal{L}_{{mse}} = \left\| z_t^s - z_t^t \right\|_2^2
\end{equation}

The total student objective combines both losses as $\mathcal{L}_s=\mathcal{L}_{pred}+\lambda \mathcal{L}_{mse}$, where $\lambda$ balances the stability of momentum-regularized prediction against the fidelity of direct teacher alignment. Through this design, the momentum-stabilized predictive mechanism ensures robust training under non-stationary teacher updates without negative samples, enabling the student to acquire a dynamics-aware terrain representation from proprioceptive history alone.

\subsection{Reward Design}
\par We maintain foundational reward terms—including 
target velocity tracking and joint torque limits which is defined in \cite{huang_learning_2025}: critical for basic 
balance control but insufficient for stair ascent tasks. Experimental 
tests indicate the model's excessive reliance on hip joints $\boldsymbol{q}^{\text{hip}}$ 
rather than knee and ankle joints during stair climbing \cite{huang_learning_2025}, which induces 
pronounced body oscillations and poor performance in physical deployments. 
To mitigate these issues, we introduce the joint range constraint reward 
$r^{\text{hr}}$ and wheelbase distance constraint reward $r^{\text{fd}}$, while augmenting 
the framework with a base position bound reward $r^{\text{pb}}$ to prevent excessive 
deviation from initial joint states and a leg balance reward $r^{\text{lb}}$ to 
promote coordinated limb motion. All reward functions and their associated 
weights are formally defined in Table \ref{tab:reward_term}, where $\boldsymbol{q}^{\text{left}}$ and $\boldsymbol{q}^{\text{right}}$ 
denote left and right leg joint position vectors, $\boldsymbol{p}^{\text{lw}}_{y}$ and $\boldsymbol{p}^{\text{rw}}_{y}$ 
represent Y-coordinates of left and right wheels in the robot frame, and 
$n^{\text{nf}}$ counts total wheel lift-off events.

\begin{table}[thpb]
\centering
\caption{Reward Terms}
\label{tab:reward_term}
\renewcommand{\arraystretch}{1.1}
\begin{tabular}{llll}
\toprule
\textbf{Reward} & \textbf{Definition} & \textbf{Weight} \\
\midrule
lin. velocity tracking & $\exp(-8.3\| \boldsymbol{v}_{xy} - \boldsymbol{v}_{xy}^{\text{cmd}} \|_2^2)$ & 7.0 \\
ang. velocity tracking & $\exp(-8.3\| \boldsymbol{\omega}_z - \boldsymbol{\omega}_z^{\text{cmd}} \|_2^2)$ & 4.0 \\
orientation & $\| \mathbf{g}_{xy} \|_2$ & -10.0 \\
Joint torques & $\| \boldsymbol{\tau} \|_2^2$ & $-1e^{-6}$ \\
dof acceleration & $\ddot{\boldsymbol{q}}^2$ & $-2.5e^{-6}$ \\
action rate & $\| \boldsymbol{a}_t - \boldsymbol{a}_{t-1} \|_2^2$ & -0.12 \\
action smoothness & $\| \boldsymbol{a}_t - 2\boldsymbol{a}_{t-1} + \boldsymbol{a}_{t-2} \|_2^2$ & -0.005 \\
feet distance & $r^{\text{fd}} = \| |\boldsymbol{p}_y^{\text{lw}} - \boldsymbol{p}_y^{\text{rw}}| - 0.28 \|_2$ & -0.01 \\
no fly & $n^{\text{nf}}$ & -0.4 \\
joint position bias & $r^{\text{pb}} = \| \boldsymbol{q} - \boldsymbol{q}^{\text{def}} \|_2$ & -1.2 \\
leg position bias & $r^{\text{lb}} = \| \boldsymbol{q}^{\text{left}} - \boldsymbol{q}^{\text{right}} \|_2$ & -0.8 \\
hip rate & $r^{\text{hr}} = \| \max(0, \boldsymbol{q}^{\text{hip}} - 0.05) \|_2$ & 0.1 \\
\bottomrule
\end{tabular}
\end{table}

Finally, the training pipeline of the DynaWM framework is summarized in Algorithm. \ref{algorithms}. where $\alpha_{ppo}$ denotes the learning rate for PPO training, and $\alpha_{wm}$ denotes the learning rate for world model optimization. $\beta$ donates the learning rate for student encoder training.

\begin{algorithm}
\caption{DynaWM Framework Training}
\label{algorithms}
\begin{algorithmic}[1]
\State Initialize environment and networks
\State Empty replay buffer ${\mathcal{D}_t}$ and ${\mathcal{D}_s}$
\State Use ${\vartheta}$ represent ${\theta^t}$, ${\theta}$ for notational brevity
\State Use $\varpi$ represent ${\theta^t}$, ${\omega}$ for notational brevity

\For{$ 0 \leq its \leq iterations $}
    \State Compute $\hat{R}_t$ and $\hat{A}_t$ using GAE
    \For{epoch:$i = 0, 1, \ldots$}
      \State World Model and Teacher Enc.update:
      \State $\varpi \gets \varpi + \alpha_{wm}\nabla_{\varpi}(\mathcal{L}_{pred}(\varpi))$ 
      \State Actor-Critic and Teacher Enc. update using PPO:
      \State $\vartheta \gets \vartheta + \alpha_{ppo}\nabla_{\theta}(L^{ppo, t}(\theta) + L^{ppo, s}(\theta))$
      \State $\phi \gets \phi + \alpha_{ppo}\nabla_{\phi}(L^{value}(\phi))$
    \EndFor

    \For{epoch:$i = 0, 1, \ldots$}
      \State Student Enc. and Student Target Enc. update:
      \State $\theta^{s} \gets \theta^{s} + \beta\nabla_{\theta^{s}}(\mathcal{L}_{{moment}}(\theta^{s}) + \lambda \mathcal{L}_{mse}(\theta^{s}))$
      \State $\hat{\theta}^s \gets \tau \hat{\theta}^s + (1-\tau \theta^s)$
    \EndFor
\EndFor
\end{algorithmic}
\end{algorithm}

\section{Experiments Result}
All training was conducted in IsaacGym \cite{2021_isaacgym} on an NVIDIA GeForce RTX 4090 GPU. We simulated 4,000 parallel environments (with 3,000 teacher and 1,000 student agents). The policy was trained at 200 Hz and updated at 50 Hz, while the hardware operated at 500 Hz to balance training speed and stability.

\subsection{training details}

\textbf{Terrain:} To improve robustness and sim-to-real transfer, training is conducted across five terrain types: flat ground, slopes ($\leq 30^\circ$), pyramid steps (step height $\leq 0.18\text{m}$), discrete uneven terrain (height variation $\leq 0.16\text{m}$), and rough surfaces (roughness $\leq 0.03\text{m}$). This diversity encourages the policy to generalize across environments and helps bridge the sim-to-real gap.

\begin{figure*}[t] 
  \centering
  \includegraphics[width=\textwidth]{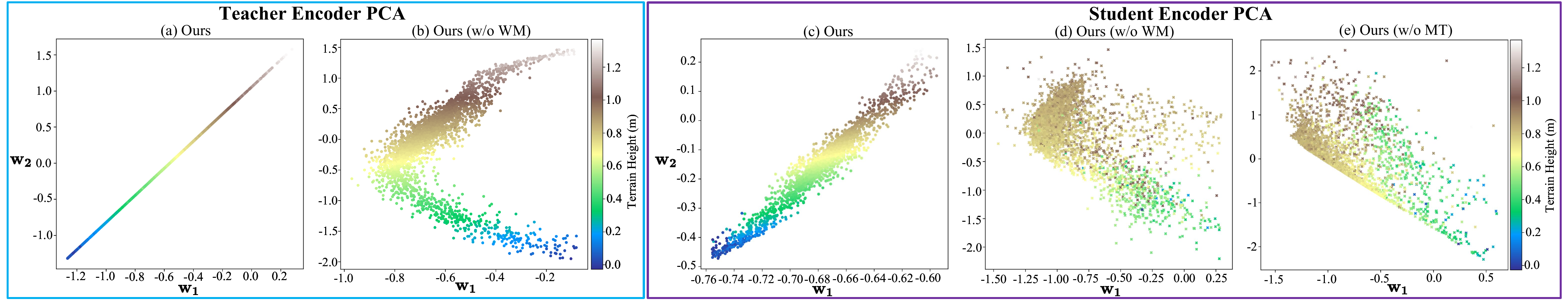} 
  \caption{\textbf{PCA visualization of learned representations.} (a) Teacher encoder with world model regularization exhibits clear terrain height stratification. (b) Teacher encoder without world model shows entangled representations. (c) Our complete student framework successfully replicates the teacher's structured manifold. (d) Student without world model suffers from dimensional collapse. (e) Student without momentum target fails to encode terrain properly.}
  \label{PCA-1}
\end{figure*}

\begin{figure*}[!t] 
\centering

  \includegraphics[width=0.98\textwidth, height=0.4\textheight, keepaspectratio]{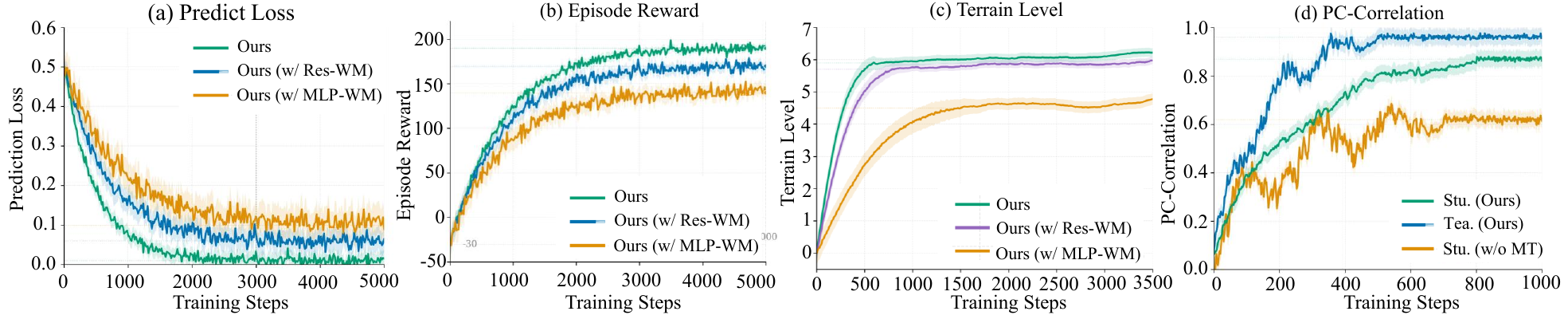}
  \caption{\textbf{Training curves comparing DynaWM against ablations.} Our Dense-LayerNorm-Swish world model achieves lower prediction loss (a), enabling more accurate future terrain state prediction and effective gradient propagation back to the encoder, which translates to superior policy learning as evidenced by higher episode reward (b) compared to ResNet and MLP variants. (c) Terrain level progression demonstrates that our method surpasses shallow MLP world model designs, where inaccurate predictions propagate detrimental gradients to the encoder and degrade training stability. (d) PC-Correlation demonstrates that momentum target stabilization enables the student to approach teacher performance, while direct distillation (w/o MT) suffers from dimensional collapse. 
  }
  \label{fig:iterations}
\end{figure*}

\textbf{Domain Randomization:} To bridge the sim-to-real gap, we applied domain randomization to key parameters during training (Table \ref{tab:domain_randomization}), including PD gains, center-of-mass offset, torque offset, and initial offsets for joint position/velocity and base pose. The robot’s initial base pose was randomized across four states—front, down, or lying on either side—with an additional pose offset applied as perturbation.

\begin{table}[t]
\centering
\caption{Domain Randomization}
\label{tab:domain_randomization}
\renewcommand{\arraystretch}{1.1}
\begin{tabular}{llll}
\toprule
\textbf{Parameter} & \textbf{Randomization Range} & \textbf{Unit} \\
\midrule
Base Mass & $\mathcal{U}(-0.1, 1.2)$ & kg \\
Base $\text{CoM}_x$ offset & $\mathcal{U}(-0.02,0.02)$ & m \\
Base $\text{CoM}_y$ offset & $\mathcal{U}(-0.01,0.01)$ & m \\
Base $\text{CoM}_z$ offset & $\mathcal{U}(-0.02,0.02)$ & m \\
Friction & $\mathcal{U}(0.1, 1.7)$ & - \\
Restitution & $\mathcal{U}(0.3, 1.0)$ & - \\
$k_{\text{p}}$ & $\mathcal{U}(0.95,1.05) \times \text{default}$ & Nm/rad \\
$k_{\text{d}}$ & $\mathcal{U}(0.98, 1.02) \times \text{default}$ & Nm/rad \\
Motor Strength & $\mathcal{U}(0.85,1.05)$ & - \\
Initial joint angle & $\mathcal{U}(0.5,1.5) \times \text{default joint angle}$ & - \\
$\boldsymbol{v}^{{cmd}}_{xy}$ & $\mathcal{U}(-0.6,0.8)$ & m/s \\
$\boldsymbol{\omega}^{{cmd}}_{z}$ & $\mathcal{U}(-0.3,0.3)$ & rad/s \\
\bottomrule
\end{tabular}
\end{table}

\subsection{Experiment Setup}
\par 1) Baseline: We conducted a comparative study of the following methods
under identical reward functions and hyperparameters:

\begin{itemize}
\item \textbf{Ours}: The complete DynaWM framework with world model regularization, momentum target encoder for stable distillation, and domain randomization.
\item \textbf{Ours (w/o WM)}: The DynaWM framework without the world model regularizer. The teacher encoder is trained solely with policy gradients, without forward-dynamics awareness, to validate the necessity of explicit terrain dynamics modeling.
\item \textbf{Ours (MLP-WM)}: The DynaWM framework with the world model implemented as a shallow MLP (4 layers) instead of the deep residual architecture, to validate the importance of network depth for dynamics prediction.
\item \textbf{Ours (Res-WM)}: The DynaWM framework with the world model using standard residual connections \cite{2017_resnet} without the specific Dense-LayerNorm-Swish block design, to isolate the contribution of the proposed architectural components.
\item \textbf{Ours (w/o MT)}: The DynaWM framework without the momentum target encoder. The student directly regresses to the online teacher representations, to validate the necessity of stable distillation targets for preventing dimensional collapse.
\item \textbf{RMA Teacher}: The teacher-only baseline from Rapid Motor Adaptation (RMA) \cite{kumar2021rma}, representing a strong privileged-information-based approach without student distillation.
\item \textbf{Concurrent Teacher-Student (CTS)}: The baseline co-training framework \cite{Wang2024CTS} that simultaneously trains teacher and student with PPO, updating the student solely via MSE-based distillation without momentum stabilization or world model regularization.
\item \textbf{Robust}: A domain randomization baseline trained without any privileged terrain information, representing the standard approach for sim-to-real transfer \cite{2018_simtoreal,2017_domain_randm}.
\end{itemize}

\par 2) Evaluation Metrics: To comprehensively evaluate the terrain encoding capability of the learned representations, we employ both qualitative visualization and evaluation metrics. We first apply Principal Component Analysis (PCA) to the latent variables $z^i_t$ collected from the encoder. PCA identifies the directions of maximum variance in the latent space by computing the eigenvectors $\mathbf{w}_1, \mathbf{w}_2, \dots$ of the covariance matrix $\mathbf{X}^\top \mathbf{X}$, where $\mathbf{X} \in \mathbb{R}^{N \times d}$is the data matrix of latent vectors. The first principal component $\mathbf{w}_1$  captures the direction of greatest variation. To investigate whether the encoder organizes its latent space by terrain height, we project each latent onto $\mathbf{w}_1$ and $\mathbf{w}_2$ and color the resulting 2D scatter plots according to the corresponding terrain height.A well-organized representation will show a smooth gradient of color along  $\mathbf{w}_1$, indicating that height is encoded along this main axis. On the other hand, a poor representation will display mixed or clustered colors without a clear pattern, suggesting that height is not effectively encoded. All metrics are evaluated on both the teacher encoder (denoted as "Tea.") and the student encoder (denoted as "Stu.") to assess representation quality at both privileged and deployment stages.

\par To further quantify this capability, we introduce three complementary metrics that assess different aspects of representation quality:

\begin{itemize}
    \item \textbf{PC-Correlation}  \cite{greenacre_principal_2022}: Measures the Pearson correlation coefficient between the scores of the first principal component and the target terrain properties (height and friction). A high absolute correlation confirms that the primary variation direction aligns with terrain features.
    \item \textbf{Linear Probing $R^2$} \cite{2017_understanding}: This approach involves training a linear regressor on top of the frozen encoder to predict terrain height or friction from the latent variables. The resulting coefficient of determination, $R^2$, measures how well these properties can be linearly decoded, providing an indication of the richness and quality of the learned representation without the need for complex nonlinear decoders.
    \item \textbf{Canonical Correlation Analysis (CCA)} \cite{2025_CCA}: Quantifies student-teacher alignment via maximum correlation between projected representation spaces. High values indicate successful distillation without dimensional collapse.
\end{itemize}

\begin{figure*}[!t] 

  \includegraphics[width=0.98\textwidth, height=0.4\textheight, keepaspectratio]{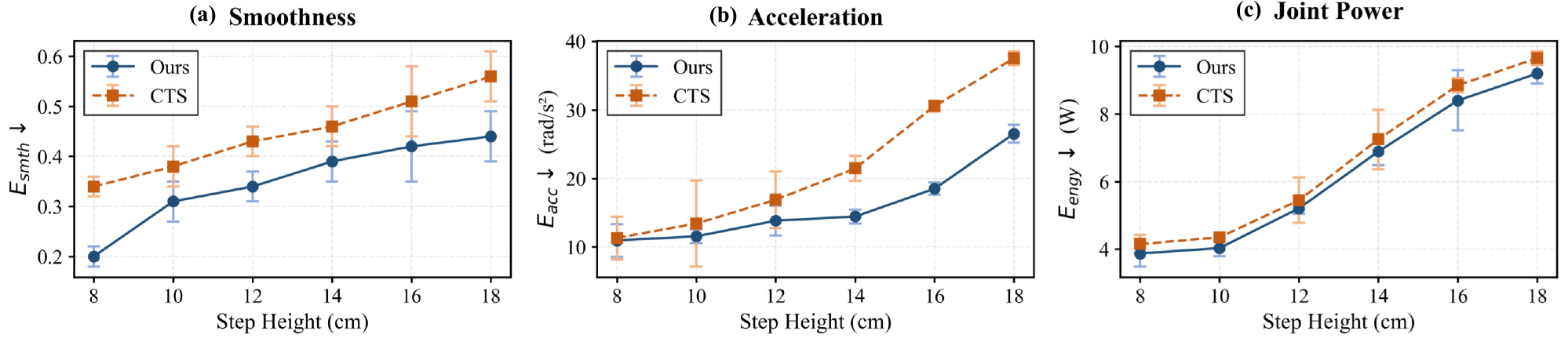}
  \caption{\textbf{Motion quality comparison against CTS across step heights.} Our method achieves superior smoothness (a) with significantly lower $E_{smooth}$, indicating reduced jerk and more fluent motion compared to CTS. The lower base acceleration (b) and comparable joint power consumption (c) demonstrate that our dynamics-aware representation enables smoother locomotion without sacrificing energy economy. $E_{smth}$, $E_{acc}$ and $E_{engy}$ is defined in \cite{huang_learning_2025}
}
  \label{fig:energy_acc_smooth}

\end{figure*}

\begin{table}[t]
\centering
\caption{Quantitative assessment of learned representations via PCA analysis}
\label{tab:pca_analysis}
\small
\begin{tabular}{@{}l@{\hspace{2pt}}lccc@{}}
\toprule
\textbf{Target} & \textbf{Encoder} & \textbf{PC-Corr.\textuparrow } & \textbf{Linear Prob.\textuparrow } & \textbf{CCA\textuparrow } \\
\midrule
\multirow{5}{*}{Height} 
 & Tea.\newline(Ours) & $\mathbf{0.96_{\pm0.01}}$ & $\mathbf{0.94_{\pm0.02}}$ & -- \\
 & Tea.\newline(w/o WM) & $0.82_{\pm0.05}$ & $0.70_{\pm0.06}$ & -- \\
 & Stu.\newline(Ours) & $0.87_{\pm0.02}$ & $0.91_{\pm0.03}$ & $\mathbf{0.89_{\pm0.03}}$ \\
 & Stu.\newline(w/o MT) & $0.35_{\pm0.08}$ & $0.32_{\pm0.09}$ & $0.42_{\pm0.07}$ \\
 & Stu.\newline(w/o WM) & $0.63_{\pm0.04}$ & $0.62_{\pm0.05}$ & $0.67_{\pm0.04}$ \\
\midrule
\multirow{5}{*}{Friction} 
 & Tea.\newline(Ours) & $\mathbf{0.91_{\pm0.03}}$ & $\mathbf{0.89_{\pm0.04}}$ & -- \\
 & Tea.\newline(w/o WM) & $0.62_{\pm0.06}$ & $0.58_{\pm0.07}$ & -- \\
 & Stu.\newline(Ours) & $0.80_{\pm0.03}$ & $0.82_{\pm0.04}$ & $\mathbf{0.83_{\pm0.04}}$ \\
 & Stu.\newline(w/o MT) & $0.28_{\pm0.09}$ & $0.25_{\pm0.10}$ & $0.35_{\pm0.08}$ \\
 & Stu.\newline(w/o WM) & $0.55_{\pm0.05}$ & $0.51_{\pm0.06}$ & $0.54_{\pm0.05}$ \\
\bottomrule
\end{tabular}
\end{table}

\begin{table}[thpb]
\centering
\footnotesize
\caption{Method Success Rates under Different Stairs Parameters}
\label{tab:success_rates}
\newlength{\colA} \setlength{\colA}{0.9cm} 
\newlength{\colB} \setlength{\colB}{0.9cm} 
\newlength{\colC} \setlength{\colC}{1.1cm} 
\begin{tabular}{@{} 
    >{\raggedright\arraybackslash}p{\colA} 
    >{\centering\arraybackslash}p{\colB} 
    *{4}{>{\centering\arraybackslash}p{\colC}} 
    @{}}
\toprule
\multicolumn{1}{r}{\multirow{4}{*}{\begin{tabular}[c]{@{}c@{}}Method \end{tabular}}} & \multicolumn{5}{c}{\textbf{Success Rate} (\%) \textuparrow } \\
\cmidrule(lr){2-6}
& \multirow{2}{=}{\centering Step Width} & \multicolumn{4}{c}{Height (cm)} \\
\cmidrule(lr){3-6}
& & 12 & 16 & 18 & \textcolor{red}{\textbf{20}} \\
\midrule
\multicolumn{1}{c}{\multirow{3}{*}{\begin{tabular}[c]{@{}c@{}} Ours \end{tabular}}} & 35cm & $98.0_{\pm1.4}$ & $96.2_{\pm2.1}$ & $84.7_{\pm3.3}$ & $83.3_{\pm3.5}$ \\
& 32cm & \textcolor{blue}{\textbf{100.0}}$_{\pm0.0}$ & $98.0_{\pm1.4}$ & \textcolor{blue}{\textbf{90.9}}$_{\pm2.6}$ & \textcolor{blue}{\textbf{87.7}$_{\pm2.9}$} \\
& \textcolor{red}{\textbf{26cm}} & \textcolor{blue}{\textbf{100.0}$_{\pm0.0}$} & \textcolor{blue}{\textbf{100.0}$_{\pm0.0}$} & \textcolor{blue}{\textbf{94.3}$_{\pm1.9}$} & \textcolor{blue}{\textbf{89.3}$_{\pm2.5}$} \\ \cline{1-6}
\addlinespace
\multicolumn{1}{c}{\multirow{3}{*}{\begin{tabular}[c]{@{}c@{}}Ours \\ (w/o WM)  \end{tabular}}} & 35cm & $98.0_{\pm1.4}$ & $90.9_{\pm2.6}$ & $83.3_{\pm3.5}$ & $76.3_{\pm4.1}$ \\
& 32cm & $98.0_{\pm1.4}$ & $96.2_{\pm2.1}$ & $84.7_{\pm3.3}$ & $73.5_{\pm4.4}$ \\
& \textcolor{red}{\textbf{26cm}} & $100.0_{\pm0.0}$ & $92.6_{\pm2.9}$ & $84.7_{\pm3.3}$ & $74.6_{\pm4.3}$ \\\cline{1-6}
\addlinespace
\multicolumn{1}{c}{\multirow{3}{*}{\begin{tabular}[c]{@{}c@{}}Ours \\ (w/o MT) \end{tabular}}} & 35cm & $96.2_{\pm2.1}$ & $90.9_{\pm2.6}$ & $84.7_{\pm3.3}$ & $64.9_{\pm5.2}$ \\
& 32cm & $98.0_{\pm1.4}$ & $83.3_{\pm3.5}$ & $82.0_{\pm3.9}$ & $73.5_{\pm4.4}$ \\
& \textcolor{red}{\textbf{26cm}} & $100.0_{\pm0.0}$ & $86.2_{\pm3.3}$ & $84.7_{\pm3.3}$ & $78.1_{\pm4.0}$ \\\cline{1-6}
\addlinespace
\multicolumn{1}{c}{\multirow{3}{*}{\begin{tabular}[c]{@{}c@{}}RMA \\ Teacher  \end{tabular}}} & 35cm & \textcolor{blue}{\textbf{100.0}$_{\pm0.0}$} & $92.6_{\pm2.9}$ & $82.0_{\pm3.9}$ & $78.1_{\pm4.0}$ \\
& 32cm & $100.0_{\pm0.0}$ & \textcolor{blue}{\textbf{98.0}$_{\pm1.4}$} & $87.7_{\pm2.9}$ & $73.5_{\pm4.4}$ \\
& \textcolor{red}{\textbf{26cm}} & $100.0_{\pm0.0}$ & $96.2_{\pm2.1}$ & $86.2_{\pm3.3}$ & $82.0_{\pm3.9}$ \\\cline{1-6}
\addlinespace
\multicolumn{1}{c}{\multirow{3}{*}{\begin{tabular}[c]{@{}c@{}}{CTS}  \end{tabular}}} & 35cm & $100.0_{\pm0.0}$ & \textcolor{blue}{\textbf{96.2}$_{\pm2.1}$} & \textcolor{blue}{\textbf{87.7}$_{\pm2.9}$} & \textcolor{blue}{\textbf{86.2}$_{\pm2.9}$} \\
& 32cm & $98.0_{\pm1.4}$ & $96.2_{\pm2.1}$ & $86.2_{\pm3.3}$ & $87.7_{\pm2.9}$ \\
& \textcolor{red}{\textbf{26cm}} & $100.0_{\pm0.0}$ & $96.2_{\pm2.1}$ & $89.3_{\pm2.5}$ & $80.6_{\pm3.5}$ \\\cline{1-6}
\addlinespace
\multicolumn{1}{c}{\multirow{3}{*}{\begin{tabular}[c]{@{}c@{}}{Robust}  \end{tabular}}} & 35cm & $100.0_{\pm0.0}$ & $89.3_{\pm3.3}$ & $84.7_{\pm3.3}$ & $72.5_{\pm4.6}$ \\
& 32cm & $98.0_{\pm1.4}$ & $94.3_{\pm2.5}$ & $86.2_{\pm3.3}$ & $73.5_{\pm4.4}$ \\
& \textcolor{red}{\textbf{26cm}} & $100.0_{\pm0.0}$ & $94.3_{\pm2.5}$ & $92.6_{\pm2.1}$ & $76.3_{\pm4.1}$ \\
\bottomrule
\end{tabular}
\\[3pt]
\footnotesize
\textcolor{red}{\textbf{Red}} indicates the zero-shot scenario with 26cm step width and 20cm height, \textcolor{blue}{\textbf{blue}} indicates the best performance achieved in each specific scenario.
\end{table}

\subsection{Training Results}

We explore the connection between prediction accuracy and the architecture of world models, as well as how this affects training dynamics and the performance of policies. Furthermore, throughout the training process, we record the variations in PC-Correlation as they relate to terrain height and friction; the final outcomes are illustrated in Fig. \ref{fig:iterations}. Following training, we assess each model five times using 150 agents per test to gauge performance across different terrains and to analyze encoder metrics related to terrain height and friction encoding. The findings are presented in Table \ref{tab:pca_analysis} and Fig. \ref{PCA-1}. The outcomes of these evaluations, which assess traversal smoothness and success rates across different step widths, are presented in Fig. \ref{fig:energy_acc_smooth} and Table \ref{tab:success_rates}. Additionally, they confirm the capacity to identify terrain features. The impact of key design choices is summarized as follows:

\textbf{World Model Enhances Terrain Encoding.} By promoting awareness of forward dynamics, the world model effectively alters the teacher encoder. In its absence, the encoder overlooks terrain characteristics that do not pertain to immediate rewards and focuses solely on optimization through policy gradients. There was a significant improvement in the PC-correlation and linear probing scores as a result of the predictive task, which required a comprehensive encoding of height and friction parameters (Fig. \ref{PCA-1}, Table \ref{tab:pca_analysis}). This linear structure reflects efficient hierarchical terrain encoding rather than collapse, as confirmed by high Linear Probing $R^2$ and success rates. Fig. \ref{fig:iterations} illustrates that the Dense-LayerNorm-Swish architecture is particularly effective in improving the accuracy of state prediction and ensuring the stability of gradient flow through deeper layers. 

\begin{figure*}[!t] 
\centering

  \includegraphics[width=0.98\textwidth, height=0.4\textheight, keepaspectratio]{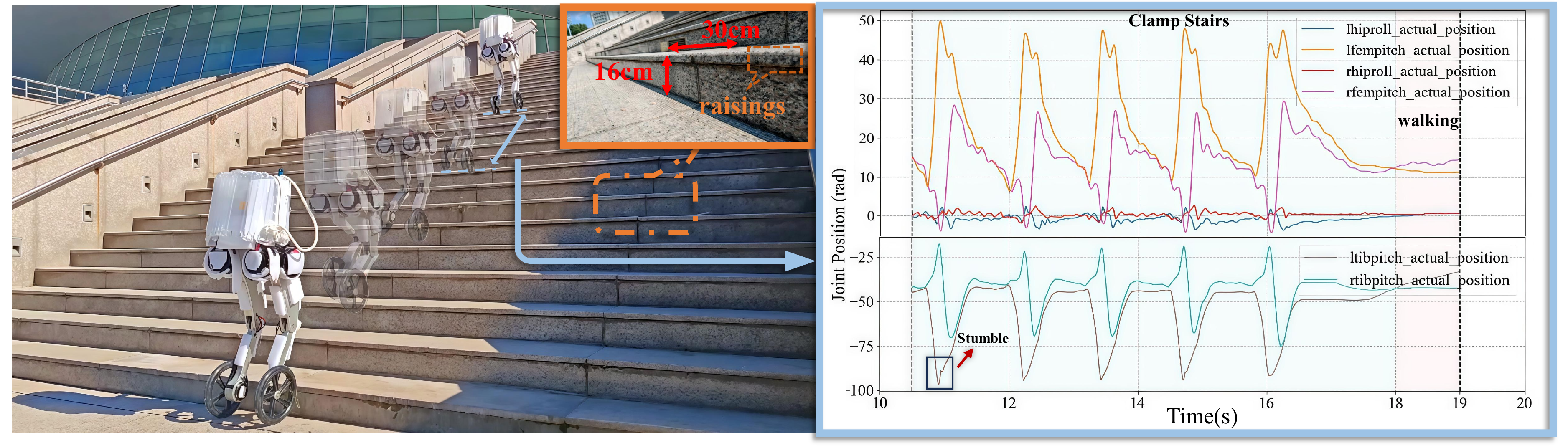}
  \caption{\textbf{The process of a bipedal robot traversing irregular continue 
  stairs.} The stairs have a step height of $16 \pm 0.8$ cm, a width of $30 \pm 0.5$ cm, 
  and a $1 \pm 0.1$ cm protrusion on the upper edge. The experimentally measured 
  joint angle trajectories during ascent verify the output stability of 
  the model.}
  \label{outdoor1}
\end{figure*}

\par \textbf{Momentum Target Encoder Prevents Dimensional Collapse.} Direct distillation from a rapidly evolving teacher leads to dimensional collapse, where the student converges to trivial representations (Fig. \ref{PCA-1}). Our momentum target encoder decouples the distillation target from the teacher's instantaneous parameters by maintaining an exponential moving average of student weights. Through BYOL training, the online student effectively develops the ability to encode terrain and maintains comprehensive representations by predicting these stable targets rather than reverting to the variable teacher (Fig. \ref{PCA-1}, Table \ref{tab:pca_analysis}). The student maintains robust encoding even during the teacher's rapid early-training changes (Fig. \ref{fig:energy_acc_smooth}).
\par \textbf{Improved Encoding Enables Smooth Locomotion.} Enhanced terrain encoding capabilities enable the policy to capture comprehensive ground contact dynamics and generate compliant motor commands. This manifests as smoother motion with lower energy consumption compared to CTS (Fig. \ref{fig:energy_acc_smooth}), and significantly improves success rates across diverse stair geometries including varying step widths and heights up to 20cm (Table \ref{tab:success_rates}). Our method achieves superior performance on challenging configurations, particularly for Wide steps at 20cm height (89.3\%) where CTS and ablations without world model or momentum target fail. The compliant approach enables reliable traversal of higher and narrower stairs where reactive control fails, validating that dynamics-aware representation learning is essential for robust locomotion in complex terrain.

\subsection{Real-World Experiment}



\begin{table}[t]
\centering
\caption{Real-world experimental results on different terrains}
\label{tab:real_world_exp}
\scriptsize
\setlength{\tabcolsep}{3pt}
\resizebox{\columnwidth}{!}{%
\begin{tabular}{@{}lcccc@{}}
\toprule
{Method} & {Height (cm)} & {Friction} & {Success Rate} & {$E_{smooth}$\textdownarrow} \\
\midrule
Ours & \multirow{2}{*}{$20_{\pm0.2}$} & \multirow{2}{*}{$0.3_{\pm0.1}$} & $\mathbf{5/5}$ & $\mathbf{0.72_{\pm0.03}}$ \\
CTS &  &  & $3/5$ & $0.91_{\pm0.05}$ \\
\midrule
Ours & \multirow{2}{*}{$16_{\pm0.8}$} & \multirow{2}{*}{$0.65_{\pm0.1}$} & $\mathbf{5/5}$ & $\mathbf{0.58_{\pm0.04}}$ \\
CTS &  &  & $4/5$ & $0.79_{\pm0.06}$ \\
\bottomrule
\end{tabular}%
}
\end{table}

\begin{figure}[thpb]
   \centering
   \includegraphics[scale=0.7]{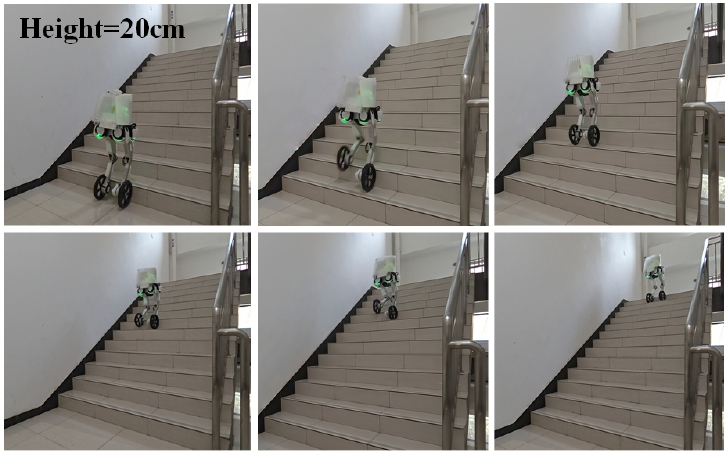}
   \caption{\textbf{Continuous stairs traversing tests} with varying step heights in unseen scenarios.}
   \label{outdoor2}
\end{figure}

We deployed the Student Encoder and Actor model trained in simulation onto our $JiaRan$ 
bipedal-wheeled robot platform. This robot possesses eight degrees 
of freedom, fulfilling all deployment requirements. All test results 
demonstrate that our approach exhibits strong robustness and stability, 
and is capable of adapting to unseen environments with a smooth motion.

To demonstrate our method's enhanced adaptability to stairs environments and the smoothness of the stairs-ascending process, we conducted five trials on two different types of unmeasured and unmodeled stair terrains. The first terrain consisted of irregular stairs featuring a $1\pm0.1$ cm protrusion on the step edge, a geometry never encountered during simulation training. The second terrain comprised 13 consecutive steps, each measuring 20 cm in height and 30 cm in width, exceeding the maximum step height seen in training by 2 cm. The robot was commanded with target velocities of 0.3 m/s and 0.2 m/s respectively, required to accelerate while overcoming disturbances induced by the irregular step edges. As shown in Table \ref{tab:real_world_exp}, our method achieved a $100\%$ success rate on both terrains, with significantly higher smoothness coefficients compared to the CTS method. This indicates that our method enables smoother and more stable stair ascent through improved understanding of unknown terrains. The final stair-ascending processes are illustrated in Fig. \ref{outdoor1} and Fig. \ref{outdoor2}, where joint position trajectories reveal highly repeatable step-ascent motions and rapid recovery from terrain disturbances, with smooth joint movements throughout and no noticeable discontinuities.

\section{CONCLUSION}

This work presents DynaWM, a dynamics-aware representation learning framework that enables smooth motion control and superior terrain adaptability for continuous stair traversal on bipedal-wheeled robots. By introducing a world model as a regularizer to enforce forward-dynamics awareness, employing a momentum target encoder to provide stable distillation targets and prevent dimensional collapse, and utilizing PCA visualization to enable transparent assessment of hierarchical terrain height encoding, our method learns dynamics-aware representations that capture comprehensive terrain geometry. Experimental validation on both simulated and real-world environments demonstrates much smoother motion control with lower base acceleration and energy consumption, alongside dependable traversal of lengthy consecutive staircases across varying step widths and heights up to 20 cm, compared to state-of-the-art baselines. Future work will explore direct integration of privileged terrain information as explicit input for real-world bipedal-wheeled robot control, leveraging exteroceptive sensing to enhance terrain adaptability across diverse and previously unseen environments.

\addtolength{\textheight}{-12cm}   









\bibliographystyle{ieeetr}

\bibliography{ref}

\end{document}